\documentclass[twoside,11pt]{article}

\usepackage{jmlr2e}

\usepackage{amsmath}
\usepackage{amssymb}
\usepackage{graphicx}
\usepackage{caption}
\usepackage{ulem} 
\usepackage{chapterbib}
\usepackage{url}
\usepackage{booktabs}

\usepackage[dvipsnames]{xcolor}

\ShortHeadings{Grounded Language Learning}{Hermann \& Hill et al.}
\firstpageno{1}

\begin{document}

\title{Grounded Language Learning in a Simulated 3D World}

\author{Karl Moritz Hermann\thanks{These authors contributed equally to this
  work.} \thanks{Corresponding authors: \href{mailto:kmh@google.com}{kmh@google.com} and \href{mailto:pblunsom@google.com}{pblunsom@google.com}.} ,
 Felix Hill\footnotemark[1] ,
Simon Green, Fumin Wang, Ryan Faulkner, Hubert Soyer, David
Szepesvari, Wojciech Marian Czarnecki, Max Jaderberg, Denis Teplyashin, Marcus Wainwright, Chris Apps,
Demis Hassabis and Phil Blunsom\footnotemark[2] \\ \\ {\it Deepmind \\ London, UK}
}

\maketitle

\begin{abstract}
We are increasingly surrounded by artificially intelligent technology that takes
decisions and executes actions on our behalf. This creates a pressing need for
general means to communicate with, instruct and guide artificial agents, with
human language the most compelling means for such communication. To achieve this
in a scalable fashion, agents must be able to relate language to the world and
to actions; that is, their understanding of language must be grounded and
embodied. However, learning grounded language is a notoriously challenging
problem in artificial intelligence research.
Here we present an agent that learns to interpret language in a simulated 3D
environment where it is rewarded for the successful execution of written
instructions.
Trained via a combination of reinforcement and unsupervised learning, and
beginning with minimal prior knowledge, the agent learns to relate linguistic
symbols to emergent perceptual representations of its physical surroundings and
to pertinent sequences of actions.
The agent's comprehension of language extends beyond its prior experience,
enabling it to apply familiar language to unfamiliar situations and to interpret
entirely novel instructions.
Moreover, the speed with which this agent learns new words increases as its
semantic knowledge grows.
This facility for generalising and bootstrapping semantic knowledge indicates
the potential of the present approach for reconciling ambiguous natural
language with the complexity of the physical world.

\end{abstract}

\section{Introduction}
Endowing machines with the ability to relate language to the physical world is a
long-standing challenge for the development of Artificial Intelligence.
As situated intelligent technology becomes ubiquitous, the development of
computational approaches to understanding grounded language has become critical
to human-AI interaction.
Beginning with \cite{twinograd1972understanding}, early attempts to ground
language understanding in a physical world were constrained by their reliance on
the laborious hard-coding of linguistic and physical rules. Modern devices with
voice control may appear more competent but suffer from the same limitation in
that their language understanding components are mostly rule-based and do not
generalise or scale beyond their programmed domains.

This work presents a novel paradigm for simulating language learning and
understanding. The approach differs from conventional computational language
learning in that the learning and understanding take place with respect to
a continuous, situated environment.
Simultaneously, we go beyond rule-based approaches to situated language
understanding as our paradigm requires agents to learn end-to-end the grounding
for linguistic expressions in the context of using language to complete tasks
given only pixel-level visual input.

The initial experiments presented in this paper take place in an extended
version of the DeepMind Lab \citep{Beattie:2016} environment, where agents are
tasked with finding and picking up objects based on a textual description of
each task.
While the paradigm outlined gives rise to a large number of possible learning
tasks, even the simple setup of object retrieval presents challenges for
conventional machine learning approaches.
Critically, we find that language learning is contingent on a
combination of reinforcement (reward-based) and unsupervised learning.
By combining these techniques, our agents learn to connect words and phrases
with emergent representations of the visible surroundings and embodied
experience. We show that the semantic knowledge acquired during this process
generalises both with respect to new situations and new language.  Our agents
exhibit zero-shot comprehension of novel instructions, and the speed at which
they acquire new words accelerates as their semantic knowledge grows.
Further, by employing a curriculum training regime, we train a single agent
to execute phrasal instructions pertaining to multiple tasks requiring distinct
action policies as well as lexical semantic and object knowledge.\footnote{See
  \url{https://youtu.be/wJjdu1bPJ04} for a video of the trained agents.}

\section{Related work}

Learning semantic grounding without prior knowledge is notoriously difficult,
given the limitless possible referents for each linguistic expression
\citep{Quine1960-QUIWO}. A learner must discover correlations in a stream of
low level inputs, relate these correlations to both its own actions and to
linguistic expressions and retain these relationships in memory. Perhaps
unsurprisingly, the few systems that attempt to learn language grounding in
artificial agents do so with respect to environments that are far simpler than
the continuous, noisy sensory experience encountered by humans
\citep{steels2008symbol,roy2002learning,krening2016explanations,Yu2017deep}.

The idea of programming computers to understand how to relate language to a
simulated physical environment was pioneered in the seminal work of
\cite{twinograd1972understanding}. His SHRDLU system was programmed to
understand user generated language containing a small number of words and
predicates, to execute corresponding actions or to ask questions requesting more
information. While initially impressive, this system required that all of the
syntax and semantics (in terms of the physical world) of each word be hard coded
a priori, and thus it was unable to learn new concepts or actions. Such
rule-based approaches to language understanding have come to be considered too
brittle to scale to the full complexities of natural language.
Since this early work, research on language grounding has taken place across a
number of disciplines, primarily in robotics, computer vision and computational
linguistics.
Research in both natural language processing and computer vision has pointed to
the importance of cross modal approaches to grounded concept learning. For
instance, it was shown that learnt concept representation spaces more faithfully
reflect human semantic intuitions if induced from information about the
perceptible properties of objects as well as from raw text
\citep{silberer2012grounded}.

Semantic parsing, as pursued the field of natural language processing, has
predominantly focussed on building a compositional mapping from natural language
to formal semantic representations that are then grounded in a database or
knowledge graph \citep{Zettlemoyer:2005:LMS:3020336.3020416,BerantCFL13}. The
focus of this direction of work is on the compositional mapping between the two
abstract modalities, natural language and logical form, where the grounding is
usually discrete and high level. This is in contrast to the work presented in
this paper where we focus on learning to ground language in low level
perception and actions.

\cite{Siskind1995} represents an early attempt to ground language in
perception by seeking to link objects and events in stick-figure animations to
language. Broadly this can be seen as a precursor to more recent work on mapping
language to actions in video and similar modalities
\citep{Siskind01,chen:icml08,YuS13}.
In a similar vein, the work of \cite{roy2002learning} applies machine learning
to aspects of grounded language learning, connecting speech or text input with
images, videos or even robotic controllers. These systems consisted of modular
pipelines in which machine learning was used to optimise individual components
while complementing hard-coded representations of the input data.
Within robotics, there has been interest in using language to facilitate
human-robot communication, as part of which it is necessary to devise mechanisms
for grounding a perceptible environment with language
\citep{hemachandra14,walter:2014:semmaps}.
In general, the amount of actual learning in these prior works is heavily
constrained, either through the extensive use of hand-written grammars and
mechanisms to support the grounding, or through simplification in terms of the
setup and environment.

Other related work focuses on language grounding from the perspective of
human-machine communication
\citep{Thomason:2015:LIN:2832415.2832516,wang2016games,ArumugamKGWT17}.
The key difference between these approaches and our work is that here again
language is grounded to highly structured environments, as opposed to the
continuous perceptible input our learning environment provides.

In the field of computer vision, image classification
\citep{krizhevsky2012imagenet} can be interpreted as aligning visual data and
semantic or lexical concepts.  Moreover, neural networks can effectively map
image or video representations from these classification networks to
human-written image captions. These mappings can also yield plausible
descriptions of visual scenes that were not observed during training
\citep{xu2015show,VendrovKFU15}. However, unlike our approach, these captioning
models typically learn visual and linguistic processing and representation from
fixed datasets as part of two separate, independent optimisations. Moreover,
they do not model the grounding of linguistic symbols in actions or a visual
stimuli that constantly change based on the exploration policy of the agent.

The idea that reinforcement-style learning could play a role in language
learning has been considered for decades \citep{chomsky1959review}. Recently,
however, RL agents controlled by deep neural nets have been trained to solve
tasks in both 2D \citep{mnih2015human} and 3D \citep{mnih2016asynchronous}
environments. Our language learning agents build on these approaches and
algorithms, but with an agent architecture and auxiliary unsupervised objectives
that are specific to our multi-modal learning task. Other recently-proposed
frameworks for interactive language learning involve unimodal (text-only)
settings \citep{narasimhan2015language,mikolov2015roadmap}.

\section{The 3D language learning environment}

To conduct our language learning experiments we integrated
a language channel into a 3D simulated world (DeepMind Lab,
  \citet{Beattie:2016}).
In this environment, an agent perceives its surroundings via a constant stream of
continuous visual input and a textual instruction. It perceives the world
 actively, controlling what it sees via
movement of its visual field and exploration of its surroundings.
One can specify the general configuration of layouts and possible
objects in this environment together with the form of language instructions that
describe how the agent can obtain rewards.  While the high-level
configuration of these simulations is customisable, the precise world
experienced by the agent is chosen at random from billions of
possibilities, corresponding to different instantiations of objects, their
colours, surface patterns, relative positions and the overall layout of the 3D
world.

To illustrate this setup, consider a very simple environment comprising two
connected rooms, each containing two
objects. To train the agent to understand simple referring expressions, the
environment could be configured to issue an instruction of the form~\emph{pick
  the $X$} in each episode. During training, the agent experiences multiple
episodes with the shape, colour and pattern of the objects themselves differing
in accordance with the instruction. Thus, when the instruction is~\emph{pick
  the pink striped ladder}, the environment might contain, in random
positions, a pink striped ladder (with positive reward), an entirely pink
ladder, a pink striped chair and a blue striped hairbrush (all with
  negative reward).

\begin{figure}
  \center
  \includegraphics[width=\textwidth]{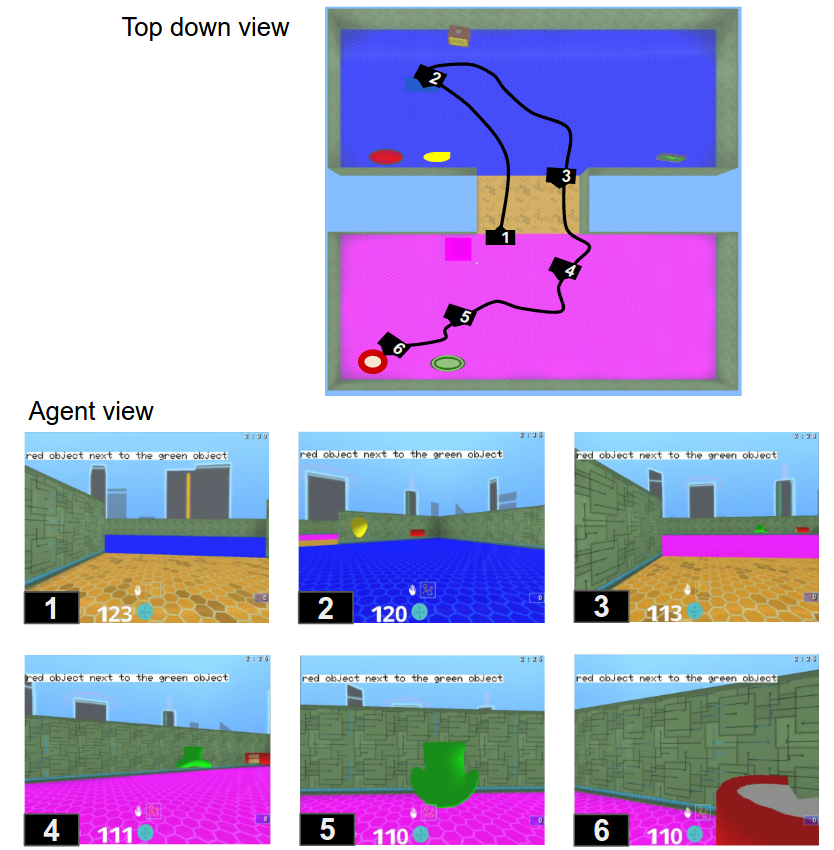}
  \caption{In this example, the agent begins in position $\mathbf{1}$ and
    immediately receives the instruction~\emph{pick the red object next to the
    green object}. It explores the two-room layout, viewing objects and their
    relative positions before retrieving the object that best conforms to the
    instruction. This exploration and selection behaviour emerges entirely from
    the reward-driven learning and is not preprogrammed. When training on a task
    such as this, there are billions of possible episodes that the agent can
    experience, containing different objects in different
    positions across different room layouts.}\label{fig:topdown}
\end{figure}

It is important to emphasise the complexity of the learning challenge faced by
the agent, even for a simple reference task such as this.
To obtain positive rewards across multiple training episodes, the agent must
learn to efficiently explore the environment and inspect candidate objects
(requiring the execution of hundreds of inter-dependent actions) while
simultaneously learning the (compositional) meanings of multi-word expressions
and how they pertain to visual features of different objects (Figure
  \ref{fig:topdown})

We also construct more
complex tasks pertaining to other characteristics of human language
understanding, such as the generalisation of linguistic predicates to novel
objects, the productive composition of words and short phrases to interpret
unfamiliar instructions and the grounding of language in relations and actions
as well as concrete objects.

\section{Agent design}

Our agent consists of four inter-connected modules optimised as a single neural
network. At each time step $t$, the visual input $v_t$ is encoded by the
convolutional~\textit{vision module} $\mathbf{V}$ and a recurrent
(LSTM, \citet{hochreiter1997long})~\textit{language module} $\mathbf{L}$ encodes the instruction string $l_t$.
A~\textit{mixing module} $\mathbf{M}$ determines how these signals are combined
before they are passed to a two-layer LSTM~\textit{action module} $\mathbf{A}$.
The hidden state $s_t$ of the upper LSTM in $\mathbf{A}$ is fed to a policy
function, which computes a probability distribution over possible motor actions
$\pi(a_t|s_t)$, and a state-value function approximator $Val(s_t)$, which
computes a scalar estimate of the agent value function for optimisation. To
learn from the scalar rewards that can be issued by the environment, the agent
employs an actor-critic algorithm \citep{mnih2016asynchronous}.

The policy $\pi$ is a distribution over a discrete set of actions. The baseline
function $Val$ estimates the expected discounted future return following the state
the agent is currently in. In other words, it approximates the state-value
function ${Val_{\pi}(s) = \mathbb{E}_{\pi}[\sum^{\infty}_{k=0} \lambda^t r_{t + k
+ 1} \mid S_t = s]}$ where $S_t$ is the state of the environment at time $t$ when
following policy $\pi$ and $r_t$ is the reward received following the action
performed at time $t$. $\lambda\in[0,1]$ is a discount parameter.

The agent's primary objective is is to
find a policy which maximizes the expected discounted return
$\mathbb{E}_{\pi}[\sum^{\infty}_{t=0} \lambda^t r_t]$.
We apply the Advantage Actor Critic algorithm~ \citep{mnih2016asynchronous} to
optimize the policy $\pi$---a Softmax multinomial distribution parametrized by
the agent's network---towards higher discounted returns.

Parameters are updated according to the RMSProp update
rule \citep{tieleman2012lecture}.  We share a single parameter vector across
$32$ asynchronous threads. This configuration offers a suitable trade-off
between increased speed and loss of accuracy due to the asynchronous
updates \citep{mnih2016asynchronous}.

\begin{figure}
\centering
  \includegraphics[height=9cm]{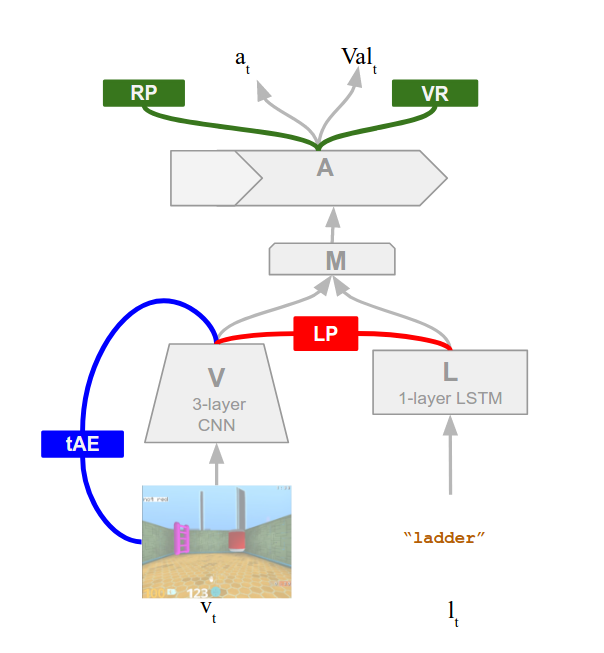}
  \caption{Schematic organisation of the
    network modules (grey) supplemented with auxiliary learning objectives
    (coloured components)}
  \label{fig:schema}
\end{figure}

Importantly, early simulation results revealed that this initial design does
not learn to solve even comparably simple tasks in our setup. As described thus
far, the agent can learn only from comparatively infrequent object selection
rewards, without exploiting the stream of potentially useful perceptual feedback
available at each time step when exploring the environment.
We address this by endowing the agent with ways to learn in an unsupervised
manner from its immediate surroundings, by means of auto-regressive objectives
that are applied concurrently with the reward-based learning and involve
predicting or modelling aspects of the agent's
surroundings \citep{jaderberg2016reinforcement}.

\textbf{Temporal autoencoding} The temporal autoencoder auxiliary
task $\mathbf{tAE}$ is designed to illicit intuitions in our
agent about how the perceptible world will change as a consequence of its actions.
 The objective is to predict the visual environment $v_{t+1}$
conditioned on the prior visual input $v_t$ and the action $a_t$ \citep{oh2015action}.
Our implementation reuses the standard visual module $\mathbf{V}$ and combines
the representation of $v_t$ with an embedded representation of $a_t$. The
combined representation is passed to a deconvolutional network to predict $v_{t+1}$.
As well as providing a means to fine-tune the visual system $\mathbf{V}$, the $\mathbf{tAE}$
auxiliary task results in additional training of the action-policy network, since
the action representations can be shared between $\mathbf{tAE}$ and the policy
network $\pi$.

\textbf{Language prediction}
To strengthen the ability of the agent to reconcile visual and linguistic modalities
 we design a word prediction objective $\mathbf{LP}$ that estimates instruction words
$l_t$ given the visual observation $v_t$, using model parameters shared with both
$\mathbf{V}$ and $\mathbf{L}$. The $\mathbf{LP}$ network can also serve to make
the behaviour of trained agents more interpretable, as the agent emits words
that it considers to best describe what it is currently observing.

The $\mathbf{tAE}$ and $\mathbf{LP}$ auxiliary networks were optimised with
mini-batch gradient descent based on the mean-squared error and
negative-log-likelihood respectively. We also experimented with reward
prediction ($\mathbf{RP}$) and value replay ($\mathbf{VR}$) as additional
auxiliary tasks to stabilise reinforcement based
training \citep{jaderberg2016reinforcement}.

Figure \ref{fig:schema} gives a schematic organisation of the agent with all the
above auxiliary learning objectives. Precise implementation details of the agent
are given in Appendix~\ref{App:C}.

\section{Experiments}

In evaluating the agent, we constructed tasks designed to test its capacity to
cope with various challenges inherent in language learning and understanding.
We first test its ability to efficiently acquire a varied vocabulary of words
pertaining to physically observable aspects of the environment. We then examine
whether the agent can combine this lexical knowledge to interpret both familiar and
unfamiliar word combinations (phrases).  This analysis includes phrases whose
meaning is dependent of word order, and cases in which the agent must induce and
re-use lexical knowledge directly from (potentially ambiguous) phrases. Finally,
we test the agent's ability to learn less concrete aspects of language,
including instructions referring to relational concepts~\citep{doumas2008theory}
and phrases referring to actions and behaviours.

\subsection{Role of unsupervised learning}

\begin{figure}[h]
\centering
  \includegraphics[height=10cm]{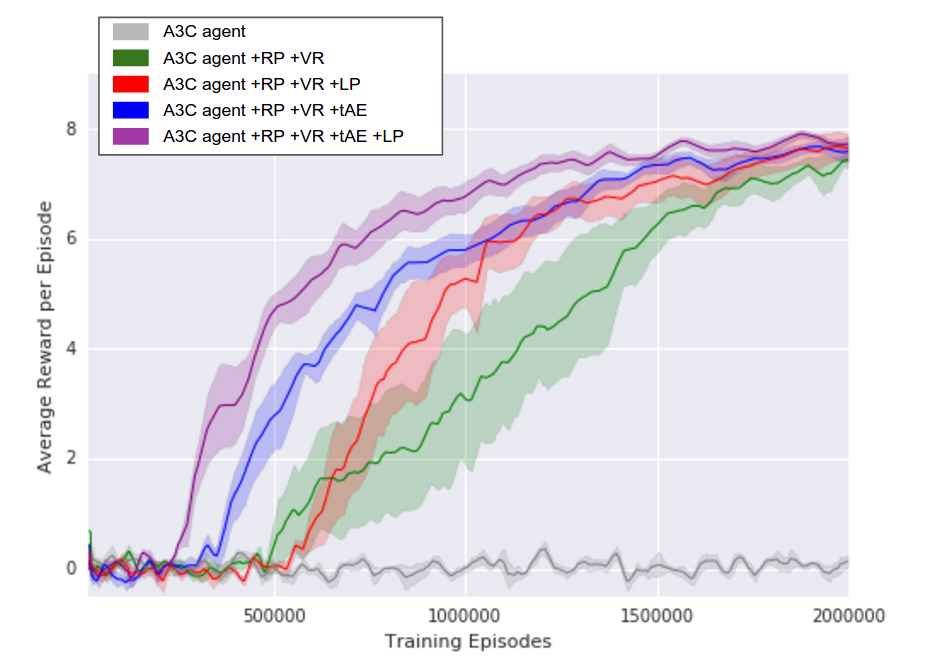}
  \caption{\textbf{Unsupervised learning via auxiliary prediction objectives
    facilitates word learning}. Learning curves for a vocabulary
    acquisition task. The agent is situated in a single room faced with two objects
    and must select the object that correctly matches the textual instruction.
    A total of 59 different words were used as instructions during training,
    referring to either the shape, colours, relative size (larger,
      smaller), relative shade (lighter, darker) or surface pattern (striped,
    spotted, etc.) of the target object.
    $\mathbf{RP}$: reward prediction, $\mathbf{VR}$: value replay,
    $\mathbf{LP}$: language prediction, $\mathbf{tAE}$: temporal autoencoder. Data
    show mean and confidence bands (CB) across best five of 16 hyperparameter
    settings sampled at random from ranges specified in the appendix.
  Training episodes counts individual levels seen during training.}
  \label{fig:auxcomparison}
\end{figure}

Our first experiment explored the effect of the auxiliary objectives on the
ability of the agent to acquire a vocabulary of different concrete words (and
associated lexical concepts). Training consisted of multiple episodes in a
single room containing two objects. For each episode, at time $t=0$, the agent
was spawned in a position equidistant from the two objects, and received a
single-word instruction that unambiguously referred to one of the two objects.
It received a reward of $1$ if it walked over to and selected the correct referent object and $-1$
if it picked the incorrect object. A new episode began immediately after an
object was selected, or if the agent had not selected either object after 300
steps.  Objects and instructions were sampled at random from the full set of
factors available in the simulation environment.\footnote{See
Appendix~\ref{App:A} for a complete list.} We trained 16 replicas for each agent
configuration (Figure~\ref{fig:auxcomparison}) with fixed hyperparameters from
the standard settings and random hyperparameters sampled uniformly from the
standard ranges.\footnote{See Appendix~\ref{App:B} for details.}

As shown in Figure~\ref{fig:auxcomparison}, when relying on reinforcement
learning alone, the agent exhibited no learning even after millions of training
episodes. The fastest learning was exhibited by an agent applying both temporal
auto-encoding and language prediction in conjunction with value replay and
reward prediction.
These results demonstrate that auto-regressive objectives can extract
information that is critical for language learning from the perceptible
environment, even when explicit reinforcement is not available.

\subsection{Word learning speed experiment}

\begin{figure}
  \includegraphics[width=\textwidth]{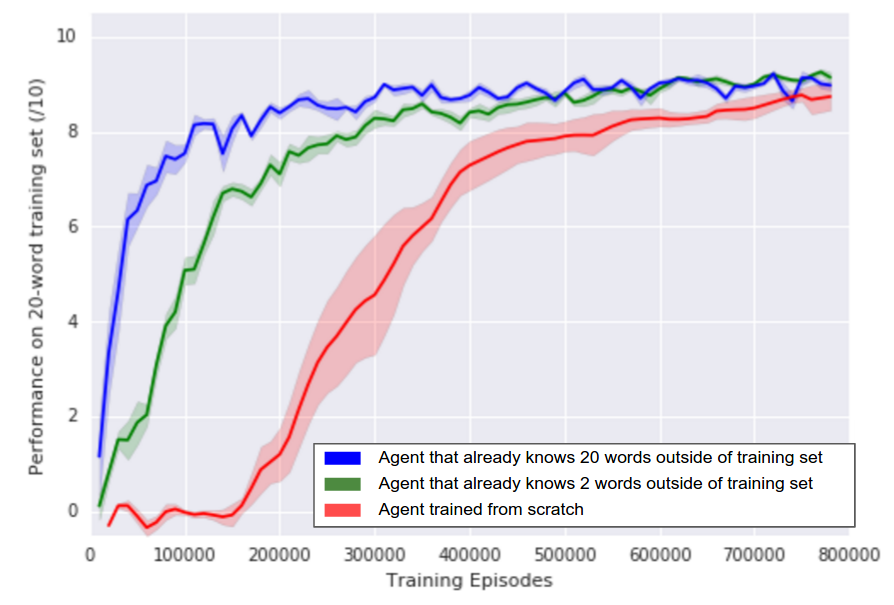}
  \caption{
    \textbf{Word learning is much faster once some words are already known} The
    rate at which agents learned a vocabulary of 20 shape words was
    measured in agents in three conditions. In one condition, the agent had
    prior knowledge of 20 shapes and their names outside of the training data
    used here. In the
    second condition, the agent had prior knowledge of two shape words outside
    of the target vocabulary (same number of pre-training steps). In the third
    condition, the agent was trained from scratch. All agents used
    $\mathbf{RP}$, $\mathbf{VR}$, $\mathbf{LP}$ and $\mathbf{tAE}$ auxiliary
    objectives. Data show mean and confidence bands across best five of 16
    hyperparameter settings in each condition, sampled at random from ranges
    specified in Appendix \ref{App:B}.
  }\label{fig:twenty}
\end{figure}

Before it can exhibit any lexical knowledge, the agent must learn various skills
and capacities that are independent of the specifics of any particular language
instruction.  These include an awareness of objects as distinct from floors or
walls; some capacity to sense ways in which those objects differ; and the ability
to both look and move in the same direction.  In addition, the agent must infer
that solving the solution to tasks is always contingent on both visual and
linguistic input, without any prior programming or explicit teaching of the
importance of inter-modal interaction. Given the complexity of this learning
challenge, it is perhaps unsurprising that the agent requires thousands of
training episodes before evidence of word learning emerges.

To establish the importance of this `pre-linguistic' learning, we
 compared the speed of vocabulary acquisition in agents with different degrees of prior
knowledge. The training set consisted of
instructions (and corresponding environments) from the twenty shape terms
\textit{banana, cherries, cow, flower, fork, fridge, hammer, jug, knife,
pig, pincer, plant, saxophone, shoe, spoon, tennis-racket, tomato, tree,
wine-glass} and~\textit{zebra.} The agent with most prior knowledge was trained
in advance (in a single room setting with two objects) on the remaining twenty
shapes from the full environment. The agent with minimal prior knowledge was
trained only on the two terms~\textit{ball} and~\textit{tv}. Both regimes of
advanced training were stopped once the agent reached an average reward of
$9.5/10$ across 1,000 episodes.  The agent with no prior knowledge
began learning directly on the training set.

The comparison presented in Figure~\ref{fig:twenty} demonstrates that much of the
initial learning in an agent trained from scratch involves acquiring visual and
motor, rather than expressly linguistic, capabilities. An agent already knowing
two words (and therefore exhibiting rudimentary motor and visual skills) learned
new words at a notably faster rate than an agent trained from scratch.
Moreover, the speed of word learning appeared to accelerate as more words were
learned. This shows that the acquisition of new words is supported not only by
general-purpose motor-skills and perception, but also existing lexical or
semantic knowledge. In other words, the agent is able to bootstrap its existing
semantic knowledge to enable the acquisition of new semantic knowledge.

\subsection{One-shot learning experiments}

\begin{figure} \begin{minipage}[b]{\textwidth}
\includegraphics[width=\textwidth]{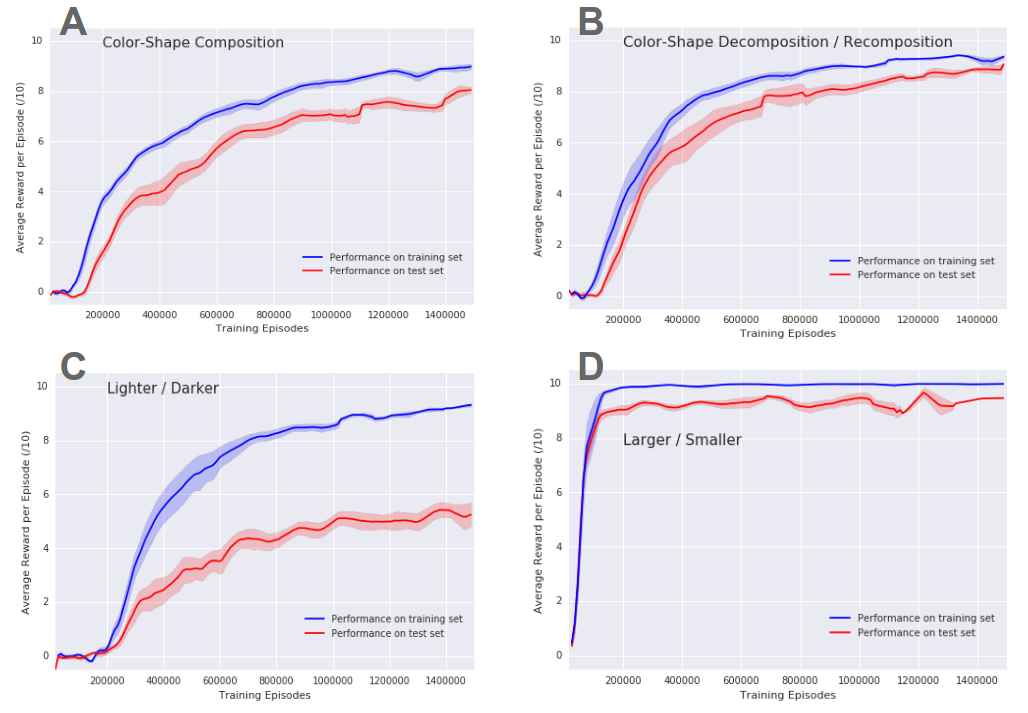}\end{minipage}
\caption{
  \textbf{Semantic knowledge generalises to unfamiliar language and objects.}
  \textbf{Composition~(A)}: training covered all shape and colour
  unigrams and $\sim90\%$ of possible colour-shape bigrams, such as~\emph{blue
  ladder}. Agents were periodically tested on the remaining 10\% of bigrams without
  updating parameters.
  \textbf{Decomposition-composition (B)}: the same regime as in A, but without
  any training on unigram descriptors.
  \textbf{Lighter / darker (C)}: agents were trained to interpret the
  terms \emph{lighter} and \emph{darker} applied to a set of colours, and tested
  on the terms in the context of a set of different colours.
  \textbf{Relative size (D)}: agents were trained to interpret the
  terms \emph{larger} and \emph{smaller} applied to a set of shapes, and tested
  on the terms in the context of a set of different shapes.
  Data show mean and CB across best five of 16 randomly sampled hyperparameter
  settings in each condition. See Appendix \ref{App:A} for hyperparameter ranges and exact
  train/test stimuli.
}\label{fig:generalisation}
\end{figure}

Two important facets of natural language understanding are the ability
to compose the meanings of known words to interpret otherwise unfamiliar phrases, and
the ability to generalise linguistic knowledge learned in one setting to make sense
of new situations.
To examine these capacities in our agent,
 we trained it in settings where its (linguistic or visual) experience was
constrained to a training set, and simultaneously as it learned from the training set,
tested the performance of the agent on situations outside of this set
(Figure~\ref{fig:generalisation}).

In the~\textbf{colour-shape composition} experiment, the training instructions were
either unigrams or bigrams. Possible unigrams were the 40 shape and the 13
colour terms listed in Appendix \ref{App:A}. The possible bigrams were any colour-shape combination
except those containing the shapes \textit{ice\_lolly, ladder, mug, pencil,
suitcase} or the colours \textit{red, magenta, grey, purple} (subsets selected
randomly). The test instructions consisted of all possible bigrams excluded from the
training set. In each training episode, the target object was rendered to match the
instruction (in colour, shape or both) and the confounding object did not
correspond to any of the bigrams in the test set. Similarly, in each test
episode, both the target object and the confounding object corresponded to
bigrams in the test instructions. These constraints ensured that the agent could
not interpret test instructions by excluding other objects or terms that it had
seen in the training set.

The~\textbf{colour-shape decomposition / composition} experiment is similar in
design to the colour-shape composition experiment. The test tasks were
identical, but the possible training instructions consisted only of the bigram
instructions from the colour-shape composition training set. To achieve above
chance performance on the test set, the agent must therefore isolate aspects of
the world that correspond to each of the constituent words in the bigram
instructions (decomposition), and then build an interpretation of novel bigrams
using these constituent concepts.

The~\textbf{relative size} and ~\textbf{relative shade} experiments were
designed to test the generality of agents' representation of relational concepts
(in this case~\textit{larger}, ~\textit{smaller},~\textit{larger}
and~\textit{darker}.  Training and testing episodes again took place in a
single room with two objects. The relative size experiment involved the 16
shapes in our environment whose size could be varied while preserving their
shape.  The possible instructions in both training and test episodes were simply
the unigrams ~\textit{larger} and~\textit{smaller}. The agent was required to
choose between two objects of the same shape but different size (and possibly
different colour) according to the instruction. All training episodes involved
target and confounding objects whose shape was either a \textit{tv, ball,
balloon, cake, can, cassette, chair, guitar, hairbrush} or \textit{hat}. All
test episodes involved objects whose shape was either an \textit{ice\_lolly,
ladder, mug, pencil} or \textit{toothbrush}.

The relative shade experiment followed the same design, but the agent
was presented with two objects of possibly differing shape that differed
only in the shade of their colouring (e.g. one light blue and one dark blue).
 The training colours were \textit{green, blue, cyan, yellow, pink, brown} and \textit{orange}.
The test colours were \textit{red, magenta, grey} and \textit{purple}.

When trained on colour and shape unigrams together with a limited number of
colour-shape bigrams, the agent naturally understood additional colour-shape
bigrams if it is familiar with both constituent words. Moreover, this ability to
productively compose known words to interpret novel phrases was not contingent
on explicit training of those words in isolation. When exposed only to bigram
phrases during training, the agent inferred the constituent lexical concepts and
reapplied these concepts to novel combinations at test time. Indeed, in this
condition (the decomposition/composition case), the agent learned to generalise
after fewer training instances than in the apparently simpler composition case.
This can be explained by by the fact that episodes involving bigram instructions
convey greater information content, such that the latter condition avails the
agent of more information per training episode. Critically, the agent's ability
to decompose phrases into constituent (emergent) lexical concepts reflects an
ability that may be essential for human-like language learning in naturalistic
environments, since linguistic stimuli rarely contain words in isolation.

Another key requirement for linguistic generalisation is the ability to extend
category terms beyond the specific exemplars from which those concepts were
learned \citep{quinn1993evidence,rogers2004semantic}. This capacity was also
observed in our agent; when trained on the relational concepts~\textit{larger}
and \textit{smaller} in the context of particular shapes it naturally applied
them to novel shapes with almost perfect accuracy. In contrast, the ability to
generalise ~\textit{lighter} and~\textit{darker} to unfamiliar colours was
significantly above chance but less than perfect. This may be because it is
particularly difficult to infer the mapping corresponding to lighter and darker
(as understood by humans) in an RGB colour space from the small number of
examples observed during training.

Taken together, these instances of generalisation demonstrate that our agent
does not simply ground language in hard coded features of the environment such
as pixel activations or specific action sequences, but rather learns to ground
meaning in more abstract semantic representations. More practically, these
results also suggest how artificial agents that are necessarily exposed to
finite training regimes may ultimately come to exhibit the productivity
characteristic of human language understanding.

\subsection{Extending learning via a curriculum}

\begin{figure}
  \includegraphics[width=\textwidth]{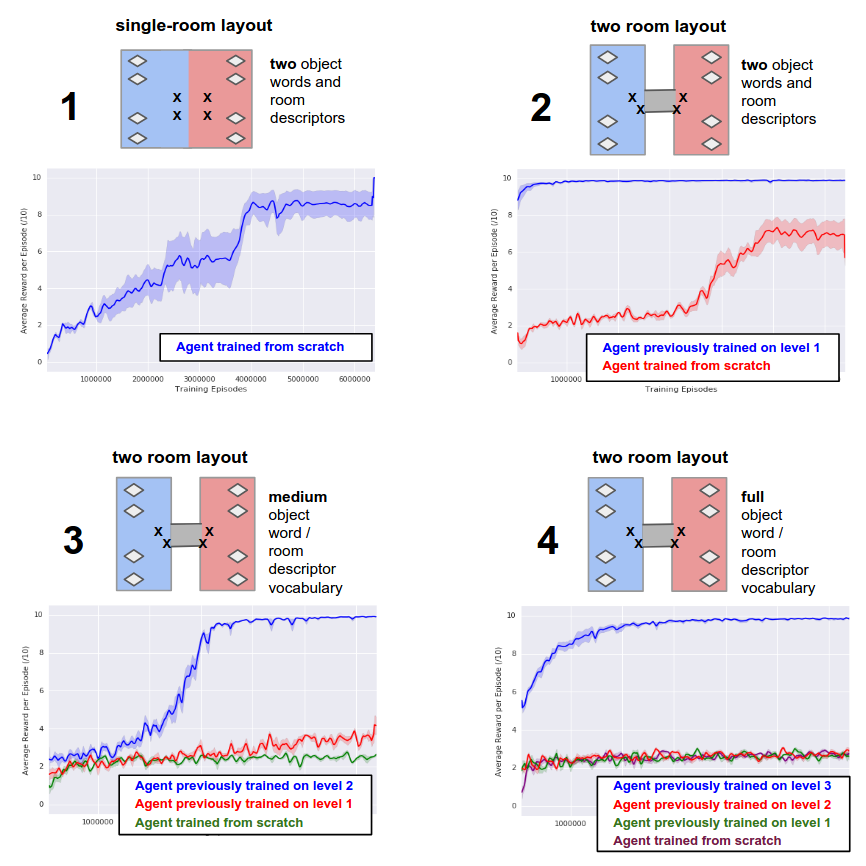}
  \caption{
    \textbf{Curriculum learning is necessary for solving more complex tasks.}
    For the agent to learn to retrieve an object in a particular room as instructed, a
    four-lesson training curriculum was required. Each lesson involved
    a more complex layout or a wider selection of objects and words, and
    was only solved by an agent that had successfully solved the previous lesson.
    The schematic layout and vocabulary scope
    for each lesson is shown above the training curves for that lesson.
   The initial (spawn) position of this agent varies randomly during
    training among the locations marked~\textbf{x}, as do the position of the four
    possible objects among the positions marked with a white diamond.
    Data show mean and CB across best five of 16 randomly sampled hyperparameter
    settings in each condition.
}\label{fig:curriculum}
\end{figure}

A consequence of the agent's facility for re-using its acquired knowledge for
further learning is the potential to train the agent on more complex language
and tasks via exposure to a curriculum of levels. Figure \ref{fig:curriculum}
shows an example for the successful application of such a curriculum, here
applied to the task of selecting an object based on the floor colour of the room
it is located in.

\begin{figure}
  \includegraphics[width=\textwidth]{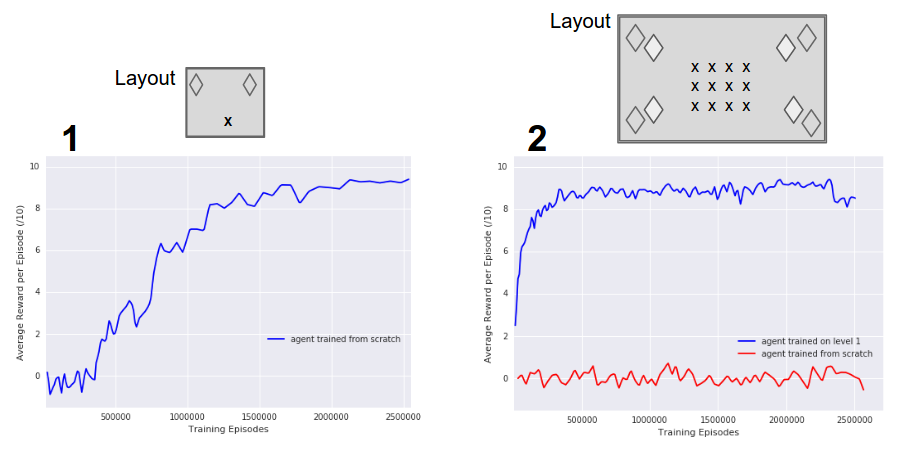}
  \caption{
     {\bf Learning curve for the referring expression agent.} The trained
    agent is able to select the correct object in a two-object setup when
    described using a compositional expression. This ability
    transfers to more complex environments with a larger number of confounding
     objects.
   }
   \label{fig:referring}
 \end{figure}

We also applied a curriculum to train an agent on a range of multi-word referring
instructions of the form \emph{pick the $X$}, where $X$ represents a string
consisting of either a single noun (shape term, such as~\textit{chair}) an
adjective and a noun (a colour term, pattern term or shade term, followed by a
shape term, such as \textit{striped ladder}) or two adjectives and a noun (a
shade term or a pattern term, followed by a colour term, followed by a shape
term, such as \textit{dark purple toothbrush}). The latter two cases were also
possible with the generic term `object' in place of a shape term. In each case,
the training episode involved one object that coincided with the instruction and
some number of distractors that did not. Learning curves for this `referring
expression agent' are illustrated in Figure \ref{fig:referring}.

\subsection{Multi-task learning}

Language is typically used to refer to actions and behaviours as much as to
objects and entities. To test the ability of our agents to ground such words in
corresponding procedures, we trained a single agent to follow instructions
pertaining to three dissociable tasks. We constructed these tasks using a
two-room world with both floor colourings and object properties sampled at
random.

\begin{figure}
  \includegraphics[width=\textwidth]{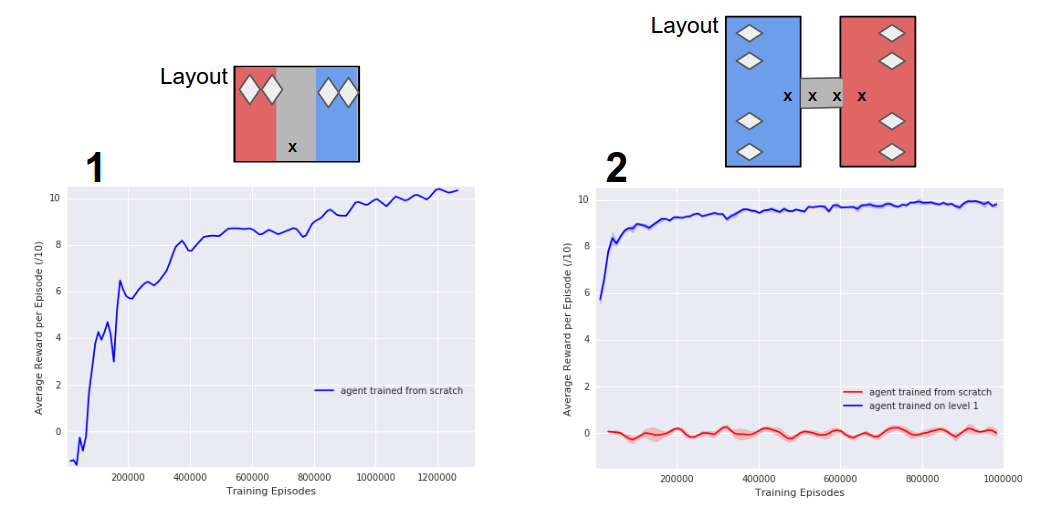}
  \caption{
    \textbf{Multi-task learning via an efficient curriculum of two steps.}
    A single agent can learn to solve a number of different tasks following a
    two-lesson training curriculum. The different tasks cannot be distinguished
    based on visual information alone, but require the agent to use the language
    input to identify the task in question.
}\label{fig:multitask}
\end{figure}

In this environment, the \textbf{Selection} task involved instructions of the
form~\emph{pick the $X$ object} or~\emph{pick all $X$}, where $X$ denotes a
colour term. The \textbf{Next to} task involved instructions of the
form~\emph{pick the $X$ object next to the $Y$ object}, where $X$ and $Y$ refer
to objects. Finally, the \textbf{In room} task involved
instructions of the form~\emph{pick the $X$ in the $Y$ room}, where $Y$ referred
to the colour of the floor in the target room.
Both the \textbf{Next to} and the \textbf{In room} task employed large degrees
of ambiguity, i.e. a given \textbf{Next to} level may contain several objects
$X$ and $Y$, but in a constellation that only one $X$ would be located next
to a $Y$.

The agent was exposed to
instances of each task with equal probability during training. The possible
values for variables $X$ and $Y$ in these instructions were~\textit{red, blue,
green, yellow, cyan} and~\textit{magenta}. The shape of all objects in the
environment was selected randomly from 40 possibilities.

As previously, a curriculum was required to achieve the best possible agent
performance on these tasks (see Figure \ref{fig:multitask}). When trained from
scratch, the agent learned to solve all three types of task in a single room
where the colour of the floor was used as a proxy for a different room. However,
it was unable to achieve the same learning in a larger layout with two distinct
rooms separated by a corridor.  When the agent trained in a single room was
transferred to the larger environment, it continued learning and eventually was
able to solve the more difficult task.\footnote{See
\url{https://youtu.be/wJjdu1bPJ04} for a video of the final trained agent.}

By learning these tasks, this agent demonstrates an ability to ground language
referring not only to single (concrete) objects, but also to (more abstract)
sequences of actions, plans and inter-entity relationships. Moreover, in
mastering the \textbf{Next to} and \textbf{In room} tasks, the agent
exhibits sensitivity to a critical facet of many natural languages, namely the
dependence of utterance meaning on word order.
The ability to solve more complex tasks by curriculum training emphasises
the generality of the emergent semantic representations acquired by the agent,
allowing it to transfer learning from one scenario to a related but more complex
environment.

\section{Conclusion}

An artificial agent capable of relating natural languages to the physical world
would transform everyday interactions between humans and technology. We have
taken an important step towards this goal by describing an agent that learns to
execute a large number of multi-word instructions in a simulated
three-dimensional world, with no pre-programming or hard-coded knowledge.  The
agent learns simple language by making predictions about the world in which that
language occurs, and by discovering which combinations of words, perceptual cues
and action decisions result in positive outcomes. Its knowledge is distributed
across language, vision and policy networks, and pertains to modifiers,
relational concepts and actions, as well as concrete objects. Its semantic
representations enable the agent to productively interpret novel word
combinations, to apply known relations and modifiers to unfamiliar objects and
to re-use knowledge pertinent to the concepts it already has in the process of
acquiring new concepts.

While our simulations focus on language, the outcomes are relevant to machine
learning in a more general sense. In particular, the agent exhibits active,
multi-modal concept induction, the ability to transfer its learning and apply
its knowledge representations in unfamiliar settings, a facility for learning
multiple, distinct tasks, and the effective synthesis of unsupervised and
reinforcement learning.
At the same time, learning in the agent reflects various effects that are
characteristic of human development, such as rapidly accelerating rates of
vocabulary growth, the ability to learn from both rewarded interactions and
predictions about the world, a natural tendency to generalise and re-use
semantic knowledge, and improved outcomes when learning is moderated by
curricula \citep{vosniadou1992mental,smith1996naming,pinker1987bootstrapping,pinker2009language}.
Taken together, these contributions open many avenues for future investigations
of language learning, and learning more generally, in both humans and artificial
agents.

\vskip 0.2in
\bibliography{../bibliography}
\appendix

\section{Agent details}
\label{App:C}
\subsection{Agent core}

At every time-step $t$ the vision module~\textbf{V} receives an $84 \times 84$
pixel RGB representation of the agent's (first person) view of the environment
($x^v_{t} \in \mathbb{R}^{3\times 84\times 84}$), which is then processed with a
three-layer convolutional neural network \citep{lecun1989backpropagation} to emit
an output representation $v_t \in \mathbb{R}^{64\times 7\times 7}$. The first
layer of the convolutional network contains 8 kernels applied at stride width 4,
resulting in 32 ($ 20 \times 20 $) output channels.  The second layer applies 4
kernels at stride with 2 yielding 64 ($ 9 \times 9 $) output channels. The third
layer applies 3 kernels at stride width 1 resulting again in 64 ($ 7 \times 7 $)
output channels.

The language module receives an input $x^l_{t} \in \mathbb{N}^s$, where $s$ is
the maximum instruction length with words represented as indices in a
dictionary.
For tasks that require sensitivity to the order of words in the language
instruction, the language module~\textbf{L} encodes $x^l_{t}$ with a recurrent (LSTM)
architecture \citep{hochreiter1997long}. For other tasks, we applied a simpler
bag-of-words (BOW) encoder, in which an instruction is represented as the sum of
the embeddings of its constituent words, as this resulted in faster training.
Both the LSTM and BOW encoders use word embeddings of dimension 128, and the
hidden layer of the LSTM is also of dimension 128, resulting in both cases in an
output representation $l_t \in \mathbb{R}^{128}$.

In the mixing module~\textbf{M}, outputs $v_t$ and $l_t$ are combined by
flattening $v_t$ into a single vector and concatenating the two resultant
vectors into a shared representation $m_t$. The output from~\textbf{M} at
each time-step is fed to the action
module~\textbf{A} which maintains the agent state $h_t \in \mathbb{R}^d$. $h_t$
is updated using an LSTM network combining output $m_t$ from~{\bf M} and
$h_{t-1}$ from the previous time-step. By default we set $d=256$ in all our
experiments.
\subsection{Auxiliary networks}
\paragraph{Temporal Autoencoder}
The temporal autoencoder auxiliary network~\textbf{tAE} samples sequences
containing two data points $x_i, x_{i+1}$ as well as one-shot action
representation $a_i \in \mathbb{N}^a$. It encodes
$x^v_i$ using the convolutional network defined by {\bf V} into $y \in
\mathbb{R}^{64\times 7\times 7}$. The feature representation is then transformed
using the action $a_i$,
\begin{align*}
  \hat{y} = W_{\hat{y}}\left( W_ba_i \odot W_vy \right),
\end{align*}
with $\hat{y} \in \mathbb{R}^{64\times 7\times 7}$. The weight matrix $W_b$
shares its weights with the final layer of the perceptron computing $\pi$ in the
core policy head.
The transformed visual encoding $\hat{y}$ is passed into a deconvolutional
network (mirroring the configuration of the convolutional encoder) to emit
a predicted input $w \in \mathbb{R}^{3\times 84\times 84}$.
The \textbf{tAE} module is optimised on the mean-squared loss between $w$ and
$x^v_{i+1}$.

\paragraph{Language Prediction}
At each time-step $t$, the language prediction auxiliary network~\textbf{LP}
applies a replica of~\textbf{V} (with shared weights) to encode $v_t$. A linear
layer followed by a rectified linear activation function is applied to transform
this representation from size $64 \times 7 \times 7 $ to a flat vector of
dimension 128 (the same size as the word embedding dimension in~\textbf{L}).
This representation is then transformed to an output layer with the same number
of units as the agent's vocabulary. The weights in this final layer are shared
with the initial layer (word embedding) weights from~\textbf{L}. The output
activations are fed through a Softmax activation function to yield a probability
distribution over words in the vocabulary, and the negative log likelihood of
the instruction word $l_t$ is computed as the loss. Note that this objective
requires a single meaningful word to be extracted from the instruction as the
target.

\section{Environment details}
\label{App:A}
The environment can contain any number of rooms connected through corridors.
A level in the simulated 3D world is described by a map (a combination of rooms),
object specifiers, language and a reward function.
Objects in the world are drawn from a fixed inventory and can be described using
a combination of five factors.

\begin{description}
  \item[Shapes (40)]   \textit{tv, ball, balloon, cake, can, cassette, chair,
    guitar, hairbrush, hat, ice\_lolly, ladder, mug, pencil, suitcase,
    toothbrush, key, bottle, car, cherries, fork, fridge, hammer, knife, spoon,
    apple, banana, cow, flower, jug, pig, pincer, plant, saxophone, shoe,
    tennis racket, tomato, tree, wine glass, zebra.}
  \item[Colours (13)] \textit{red ,  blue ,  white ,  grey ,  cyan ,  pink ,
    orange ,  black , green , magenta , brown , purple , yellow.}
  \item[Patterns (9)] \textit{plain, chequered, crosses, stripes, discs,
      hex, pinstripe, spots, swirls.}
  \item[Shades (3)] \textit {light, dark, neutral.}
  \item[Sizes (3)] \textit {small, large, medium.}
\end{description}

Within an environment, agent spawn points and object locations can be specified
or randomly sampled. The environment itself is subdivided into multiple rooms
which can be distinguished through randomly sampled (unique) floor colours.
We use up to seven factors to describe a particular object: the five
object-internal factors, the room it is placed in and its proximity to another
object, which can itself be described by its five internal factors.

In all simulations presented here, reward is attached to picking up a particular
object. Reward is scaled to be in $[-10;10]$ and, where possible, balanced so
that a random agent would have an expected reward of 0. This prevents agents
from learning degenerate strategies that could otherwise allow them to perform
well in a given task without needing to learn to ground the textual
instructions.

\section{Hyperparameters}
\label{App:B}
Tables 1 and 2 show parameter setting
used throughout the experiments presented in this paper.
We report results with confidence bands (CB) equivalent to $\pm$ one standard
deviation on the mean, assuming normal distribution.

\begin{table}[h]
\tiny
\begin{tabular}{@{}lll@{}}
\toprule
Hyperparameter              & Value       & Description                                                                                                                                    \\ \midrule
train\_steps                & 640m & Theoretical maximum number of time steps (across all episodes) for which the agent will be trained.                                      \\
env\_steps\_per\_core\_step & 4           & Number of time steps between each action decision (action smoothing)                                                                           \\
num\_workers                & 32          & Number of independent workers running replicas of the environment with asynchronous updating.                                              \\
unroll\_length              & 50          & Number of time steps through which error is backpropagated in the core LSTM action module                                                      \\
                            &             &                                                                                                                                                \\
\bf{auxiliary networks}         &             &                                                                                                                                                \\
vr\_batch\_size             & 1           & Aggregated time steps processed by value replay auxiliary for each weight update.        \\
rp\_batch\_size             & 10          & Aggregated time steps processed by reward prediction auxiliary for each weight update.   \\
lp\_batch\_size             & 10          & Aggregated time steps processed by language prediction auxiliary for each weight update. \\
tae\_batch\_size            & 10          & Aggregated time steps processed by temporal AE auxiliary for each weight update.         \\
                            &             &                                                                                                                                                \\
\bf{language encoder}                &             &                                                                                                                                                \\
encoder\_type               & BOW         & Whether the language encoder uses an additive bag-of-words (BOW) or an LSTM architecture.                                                      \\
                            &             &                                                                                                                                                \\
\bf{cost calculation}                        &             &                                                                                                                                                \\
additional\_discounting     & 0.99        & Discount used to compute the long-term return R\_t in the A3C objective                                                                        \\
cost\_base                  & 0.5         & Multiplicative scaling of all computed gradients on the backward pass in the network                                                           \\
                            &             &                                                                                                                                                \\
\bf{optimisation}             &             &                                                                                                                                                \\
clip\_grad\_norm            & 100         & Limit on the norm of the gradient across all agent network parameters (if above, scale down)                                                   \\
decay                       & 0.99        & Decay term in RMSprop gradient averaging function                                                                                              \\
epsilon                     & 0.1         & Epsilon term in RMSprop gradient averaging function                                                                                            \\
learning\_rate\_finish      & 0           & Learning rate at the end of training, based on which linear annealing of is applied.                                 \\
momentum                    & 0           & Momentum parameter in RMSprop gradient averaging function                                                                                      \\ \bottomrule
\end{tabular}
\label{tab:hypA}
\caption{Agent hyperparameters that are fixed throughout our experimentation
     but otherwise not specified in the text.}
\end{table}

\begin{table}[]
\tiny
\centering
\begin{tabular}{@{}lll@{}}

\toprule
Hyperparameter        & Value                     & Description                                                                                                                   \\ \midrule
\bf{auxiliary networks}   &                           &                                                                                                                               \\
vr\_weight            & \textit{uniform(0.1, 1)}           & Scalar weighting of value replay auxiliary loss relative to the core (A3C) objective.                                     \\
rp\_weight            & \textit{uniform(0.1, 1)}           & Scalar weighting of reward prediction auxiliary loss.                                \\
lp\_weight            & \textit{uniform(0.1, 1)}           & Scalar weighting of language prediction auxiliary loss. \\
tae\_weight           &\textit{uniform(0.1, 1)}           & Scalar weighting of temporal autoencoder prediction auxiliary. \\
                      &                           &                                                                                                                               \\
\bf{language encoder}          &                           &                                                                                                                               \\
embed\_init           & \textit{uniform(0.5, 1)}           & Standard deviation of normal distribution (mean = 0) for sampling \\
&& initial values of word-embedding weights in\bf{ L}.            \\
                      &                           &                                                                                                                               \\
\bf{optimisation}          &                           &                                                                                                                               \\
entropy\_cost         & \textit{uniform(0.0005, 0.005)}    & Strength of the (additive) entropy regularisation term in the A3C cost function.                                               \\
learning\_rate\_start & \textit{loguniform(0.0001, 0.002)} & Learning rate at the beginning of training \\
&& annealed linearly to reach learning\_rate\_finish at the end of train\_steps. \\ \bottomrule
\end{tabular}
\label{tab:hypB}
  \caption{Agent hyperparameters that randomly sampled in order to yield
    different replicas of our agents for training. \textit{uniform($x, y$)}
    indicates that values are sampled uniformly from the range $[x, y]$.
    \textit{loguniform($x, y$)} indicates that values are sampled from a
    uniform distribution in log-space (favouring lower values) on the
    range $[x, y]$.
  }
\end{table}

\end{document}